\documentclass[letterpaper, 10 pt, conference]{ieeeconf}
\usepackage{amsmath,amsfonts}
\usepackage{algorithmic}
\usepackage{algorithm}
\usepackage{array}
\usepackage[caption=false,font=normalsize,labelfont=sf,textfont=sf]{subfig}
\usepackage{textcomp}
\usepackage{stfloats}
\usepackage{url}
\usepackage{verbatim}
\usepackage{graphicx}
\usepackage{cite}
\usepackage{float}                  
\usepackage{subfig}               
\usepackage{overpic}               
\usepackage[most]{tcolorbox}
\usepackage{bm}
\usepackage{amsmath}
\usepackage{amsthm}
\usepackage{amsfonts}

\DeclareMathAlphabet{\mathpzc}{OT1}{pzc}{m}{it}

\DeclareFontFamily{U}{jkpmia}{}
\DeclareFontShape{U}{jkpmia}{m}{it}{<->s*jkpmia}{}
\DeclareFontShape{U}{jkpmia}{bx}{it}{<->s*jkpbmia}{}
\DeclareMathAlphabet{\mathfrak}{U}{jkpmia}{m}{it}

\newcommand{\bW}{\bm{W}}
\newcommand{\bc}{\bm{c}}
\newcommand{\bC}{\bm{C}}
\newcommand{\bS}{\bm{S}}
\newcommand{\bJ}{\bm{J}}
\newcommand{\bF}{\bm{F}}
\newcommand{\bD}{\bm{D}}
\newcommand{\q}{\bm{q}}
\renewcommand{\v}{\bm{v}}

\newcommand{\bx}{\bm{x}}
\newcommand{\bd}{\bm{d}}

\newcommand{\x}{\bm{x}}

\newcommand{\ba}{\bm{a}}
\newcommand{\bb}{\bm{b}}
\newcommand{\be}{\bm{e}}

\newcommand{\bM}{\bm{M}}

\newcommand{\bK}{\bm{K}}

\newcommand{\bP}{\bm{P}}
\newcommand{\bU}{\bm{U}}

\newcommand{\bn}{\bm{n}}

\newcommand{\bQ}{\bm{Q}}
\newcommand{\bR}{\bm{R}}

\newcommand{\y}{\bm{y}}%

\newcommand{\p}{{\bm{p}}}%
\newcommand{\e}{{\bf e}}%
\newcommand{\R}{\mathbb{R}}

\newcommand{\bA}{\bm{A}}

\newcommand{\bB}{\bm{B}}

\newcommand{\J}{\mathcal{J}}

\newcommand{\Q}{\mathcal{Q}}

\definecolor{black}{rgb}{0.3,0.3,0.3}
\definecolor{blackb}{rgb}{0.3,0.3,0.3}
\definecolor{brownb}{rgb}{0.3,0.3,0.3}

\newtcbtheorem[no counter]{Definition}{Definition}{
  enhanced,
  rounded corners,
  attach boxed title to top left={
    yshifttext=-1mm
  },
  colback=white,
  colframe=black,
  fonttitle=\bfseries,
  coltitle=white,
  boxrule=0.5pt,
  boxed title style={
    rounded corners,
    size=small,
    colback=black,
    colframe=black,
  } 
}{def}

\IEEEoverridecommandlockouts 

\title{\LARGE \bf Task-Space Riccati Feedback based Whole Body Control for Underactuated Legged Locomotion}

\author{$\text{Shunpeng Yang}^1$, $\text{Zejun Hong}^1$, $\text{Sen Li}^2$, $\text{Patrick Wensing}^3$, $\text{Wei Zhang}^1$ and $\text{Hua Chen}^{1\dagger}$
\thanks{$^1$School of System Design and Intelligent Manufacturing, Southern University of Science and Technology, China}
\thanks{$^2$Department of Civil and Environmental Engineering, The Hong Kong University of Science and Technology, China}
\thanks{$^3$Department of Aerospace and Mechanical Engineering, University of Notre Dame, USA. }}

\begin{document}

\maketitle

\begin{abstract}
This manuscript primarily aims to enhance the performance of whole-body controllers(WBC) for underactuated legged locomotion. We introduce a systematic parameter design mechanism for the floating-base feedback control within the WBC. The proposed approach involves utilizing the linearized model of unactuated dynamics to formulate a Linear Quadratic Regulator(LQR) and solving a Riccati gain while accounting for potential physical constraints through a second-order approximation of the log-barrier function. And then the user-tuned feedback gain for the floating base task is replaced by a new one constructed from the solved Riccati gain. Extensive simulations conducted in MuJoCo with a point bipedal robot, as well as real-world experiments performed on a quadruped robot, demonstrate the effectiveness of the proposed method. In the different bipedal locomotion tasks, compared with the user-tuned method, the proposed approach is at least 12\% better and up to 50\% better at linear velocity tracking, and at least 7\% better and up to 47\% better at angular velocity tracking. In the quadruped experiment, linear velocity tracking is improved by at least 3\% and angular velocity tracking is improved by at least 23\% using the proposed method.
\end{abstract}

\section{Introduction}
Legged locomotion plays a pivotal role in enabling legged robots to traverse diverse and challenging environments. In recent years, whole-body control(WBC) has become a widely accepted approach for trajectory stabilization in dynamic legged locomotion. However, its performance relies on the selection of certain parameters, such as the weighting matrices reflecting task priorities and the proportional-derivative(PD) feedback gains in task space\cite{wensing2023optimization, kim2019highly, gehring2017quadrupedal, kuindersma2016optimization}. Tailoring these parameter settings for different robots often requires a substantial time commitment and relies heavily on engineering experience. To address this problem, Bayesian optimization has been proposed to determine the value of those parameters\cite{d2022automatic, yuan2019bayesian}. Such methods work well when the parameters are fixed (such as the weight matrices in WBC), but if the required parameters themselves vary with the system state and external inputs, the method relies on new data for parameter updates, which inevitably introduces delays in the process. Using Bayesian estimation to find the optimal PD parameters assumes that the PD parameters are constant and do not change over time and system states. For underactuated multi rigid body system with floating base, such an assumption is typically violated, since such dynamics cannot be transformed into a simple second-order integrator through feedback linearization\cite{olfati2001nonlinear}. The feedback control law for such systems should be nonlinear, and linear controllers can only provide local approximations that vary globally with the system's states.

\begin{figure}
    \centering
    \includegraphics[width=0.4\textwidth]{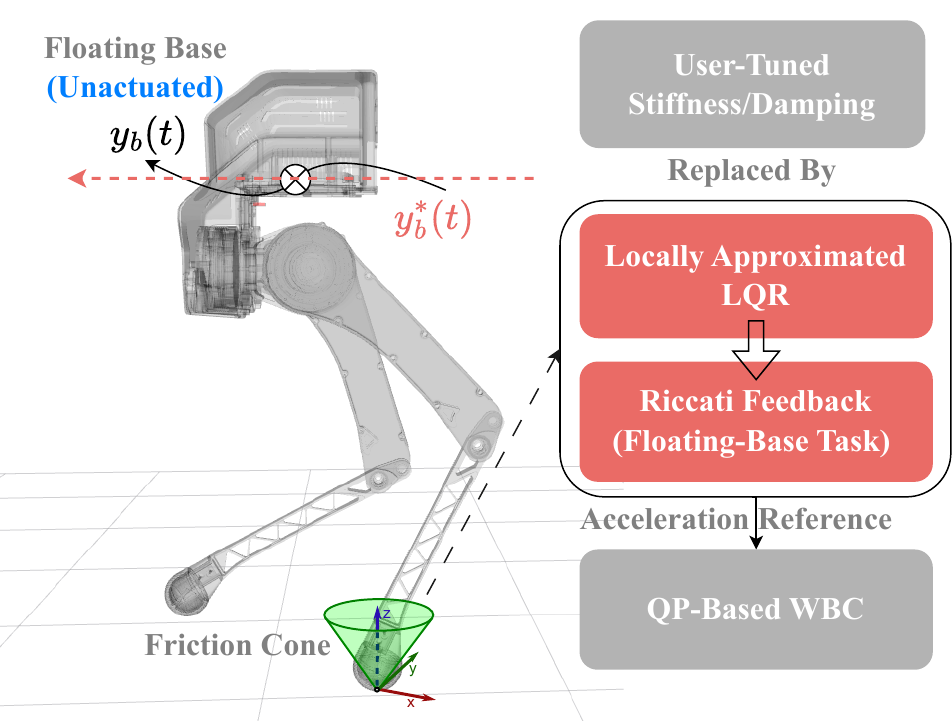}
    \vspace{-0.5cm}
    \caption{Illustration of the proposed approach for stabilizing the floating-base trajectory $\y_b(t)$ along its reference $\y_b^*(t)$. For floating-base task, the user-tuned PD feedback is repalced by a Riccati-based feedback. The Riccati gain is derived by considering the impact of friction-cone constraints and the column space of the contact Jacobian within the formualted LQR. }
    \label{fig:enter-label}
    \vspace{-0.8cm}
\end{figure}

There is another category of methods that couple the upper-level motion planner with the whole-body controller\cite{mastalli2022agile, grandia2019feedback}. In such methods, the upper-level motion planner typically yields a local feedback control law through utilizing a whole-body model and algorithms such as differential dynamic programming. This control law can be directly used in WBC and the PD feedback gains of some tasks are not necessary, such as momentum task. Hence, this particular class of methods exhibits the potential to mitigate the need for extensive parameter tuning in the context of WBC. Nonetheless, in cases where kino-dynamic model is employed in the planning, the joint velocity within this model may experience sudden fluctuations due to the absence of joint dynamics. Consequently, the integration of a method known as loop shaping becomes imperative to fortify the system's robustness in such instances. Furthermore, an alternative variation-based method has been explored in the context of two-leg balance of a quadruped robot, as discussed in \cite{9033976}. However, the controller developed in this study is not suitable for trajectory tracking due to the inherent variation around an equilibrium point. To address this limitation, an extension of this method for trajectory tracking is proposed in \cite{9560855}, where the variations are performed around an optimized trajectory. Nevertheless, it should be noted that this type of method uses centroidal dynamics and contact force is mapped to joint torque by contact Jacobian matrix without consideration of full dynamics. Consequently, its control effectiveness remains constrained.


As mentioned before, there are various parameters in WBC such as the weighting matrices reflecting task priorities and the PD gains in task space. Among them, tasks related to floating base are particularly crucial due to their close relationship with the inherent system underactuation which is known to be difficult to handle. Therefore, the core of our research lies in determining how to set the PD gains for the floating base in WBC.

In this paper, we first highlight that the conventional approach of using user-tuned stiffness and damping within the floating-base feedback fails to adequately consider system's underactuation, leading to potential performance degradation in practice. In general, a substantial amount of time is needed to meticulously tune the parameters in order to achieve desirable tracking performance.

To address this challenge, we propose an efficient method for determining the feedback gain in the floating-base task within WBC. Our approach involves the linearization of the unactuated dynamics around a trajectory reference and the construction of a LQR that explicitly incorporates the system's underactuation. The solved Riccati gain is then utilized in combination with the linear model to derive a new feedback gain for the floating-base task, rather than directly replacing user-tuned PD gains. Compared with the heuristic parameter tuning of traditional methods, our proposed method fully considers the long-term impact of system underactuation on the floating-base task, leading to more effective tracking of the floating-base motion. The superiority of our proposed approach over the user-tuned method in floating-base task is further substantiated through comprehensive experimental evaluation.

\section{Problem Description}

The dynamics of legged robot can be modeled by Newton-Euler equation
\vspace{-0.1cm}
\begin{equation}\label{eq: mrbd_dyn}
    \bM(\q)\dot\v + \bb(\q,\v) = \bS^T\bm{\tau} + \bJ_c^T(\q)\bm{\lambda}
\end{equation}
where $\q\in\Q\simeq\R^{n_v}$ is the system configuration, $\v \in \R^{n_v}$ is the generalized velocity, and $\bm{\tau}\in\R^{n_u}$ denotes its input (torque/force generated by actuator). The contact force or wrench $\bm{\lambda}\in\R^{n_c}$ satisfies the physical constraints $\mathbb{S} = \{\bm{\lambda}\in\R^{n_c}\ |\ h(\bm{\lambda})\le \bm{0}\}$ and contact points consistent with a pre-defined gait are required to maintain contact:
\vspace{-0.1cm}
\begin{equation}\label{eq: maintain_contact}
    \Phi_c(\q) = \bm{0},\ \bJ_c(\q)\v = \bm{0}
\end{equation}
and $\bJ_c(\q)$ is the Jacobian of $\Phi_c(\q)$. In the rest, we denote by $\mathbb{G}$ the time-variant set of all the states $(\q,\v)$ satisfying eq.~\eqref{eq: maintain_contact}, which is generally nonlinear and nonconvex.
\subsection{Floating-Base Task}
The floating-base pose in the task space is denoted by $\y_b = (\p,\bm{\theta}) \in\R^6$ where $\p,\bm{\theta}$ denote the position and Euler angles of floating base respectively. And the relationship between configuration $\q$ and $\y_b$ is described by $\y_b = \Phi_b(\q)$.

Given a trajectory reference $\y^*_b(t)$ of floating base, stabilizing the floating-base trajectory $\y_b(t)$ to follow the reference is one important task of WBC. One widely adopted approach for floating-base trajectory stabilization entails employing task-space linear feedback law combined with QP-based inverse dynamics. 

Now the task-space dynamics is regarded as a double integrator $\ddot \y_b = \ba_b$ where the input $\ba_b\in\R^6$ is the desired acceleration of floating base. A user-tuned linear feedback law for the double-integrator system is given as 
 \begin{equation}\label{eq: tsfb_fb}
        \ba_b = -\bK_p^b(\y_b-\y^*_b) - \bK_d^b(\dot\y_b-\dot\y_b^*) + \ddot \y_b^*
    \end{equation}
where $\bK_p^b\in\R^{6\times 6},\bK_d^b\in\R^{6\times 6}$ are chosen as positive-definite matrices. For the double-integrator system, the tracking error denoted by
\begin{equation}
    \be_b(t)= \begin{bmatrix}
    \y_b(t)-\y^*_b(t) \\
    \dot\y_b(t)-\dot\y^*_b(t)
\end{bmatrix}
\end{equation}
can converge to zero with the aforementioned choice of $\bK_p^b,\bK_d^b$. Readers are referred to\cite{kim2019highly, kuindersma2016optimization, 6631008} for more discussions about this approach. 

\subsection{QP-Based Inverse Dynamics}
There are multiple approaches to implementing WBC\cite{kim2019highly, kuindersma2016optimization}, and this paper primarily focuses on the inverse dynamics formulation based on quadratic programming (QP). The QP-based inverse dynamics at state $(\q,\v)$ is formulated as the following quadratic programming:
\begin{subequations}
    \label{opt: qp_inv_dyn}
    \begin{align}
        \min_{\dot\v, \bm{\tau},\bm{\lambda}}&\quad\Big\{\left\|\bJ_{b}(\q)\dot\v+\dot\bJ_{b}(\q)\v - \ba_b\right\|^2_{\bW} + \ell(\dot\v,\bm{\tau},\bm{\lambda})\Big\} \\
        &\text{subj. to}\quad \text{dynamics equation } \eqref{eq: mrbd_dyn},\\ &\quad\quad\quad\quad-\bm{\tau}_b \le \bm{\tau}\le\bm{\tau}_b,\ \bC\bm{\lambda}\le \bn,\label{eq: cstrs_wbc}\\
        &\quad\quad\quad\quad\frac{d}{dt}(\bJ_c(\q)\v) = \bJ_{c}(\q)\dot\v+\dot\bJ_{c}(\q)\v = \bm{0}, \label{eq: maintain_contact_cstr_wbc}
    \end{align}
\end{subequations}
where the weight matrix $\bW$ is positive definite and $\ell(\dot\v,\bm{\tau},\bm{\lambda})$, a convex quadratic form, represents other tasks for legged locomotion, such as foot swinging and input penalization. Besides, the friction-cone constraints is replaced by its inner approximation $\mathbb{\hat S} = \{\bm{\lambda}\in\R^{n_c}\ |\ \bC\bm{\lambda}\le \bn\}$ which contains some pyramids for different contact points.




\subsection{Problem}\label{sec: Problem}
Obviously, the guarantee of convergence to zero for $\be_b(t)$ relies on the condition that $\dot\v^*$ solved by the QP \eqref{opt: qp_inv_dyn} always satisfies 
\begin{equation} \label{eq: zero_cost_cond}
    \bJ_{b}(\q)\dot\v^*+\dot\bJ_{b}(\q)\v = \ba_b.
\end{equation}
In the event that this convergence condition is violated, it will indeed lead to poor tracking performance or even cause the error $\be_b(t)$ to diverge from $\bm{0}$. To alleviate such dilemma, it potentially needs much engineering effort to tune the parameters $\bK_p^b,\bK_d^b$. Even after sufficient parameter tuning, the feedback gain remains fixed in control process, which limits the performance and robustness of the controller.


Therefore, in the subsequent section, we first investigate what factors lead to the inability to solve $\dot\v^*$ from the QP \eqref{opt: qp_inv_dyn} so that it satisfies eq.~\eqref{eq: zero_cost_cond}. This analysis enables the development of a more reasonable approach to determine $\ba_b$ based on the inherent properties of the legged robot. By taking these factors into account, we come up with a more effective task-space feedback for floating base.

\section{Problem Analysis}
In this section, we will analyze why the user-tuned approach, as a model-free method, has limited effectiveness in task-space feedback for floating base based on the underactuation of the system.

According to the definition of full actuation and underactuation \cite{underactuated}, the dynamic system defined by eq.~\eqref{eq: mrbd_dyn} and eq.~\eqref{eq: maintain_contact} is fully actuated at the state $(\q,\v)$ if there exists a $\bm{\tau}$ which produces the desired response for every $\dot \v$. Otherwise, it is underactuated at state $(\q,\v)$.
And the condition $n_v \leq n_u$ represents a necessary condition for a system to possess full actuation. However, this condition is not satisfied in the case of a legged robot. Consequently, a floating-base dynamic system described by eq.~\eqref{eq: mrbd_dyn} and incorporating the contact model outlined in eq.~\eqref{eq: maintain_contact} can be classified as an underactuated system. As a result of this underactuation, the computed acceleration vector $\ba_b(t)$ derived from eq.~\eqref{eq: tsfb_fb} may fail to satisfy equation eq.~\eqref{eq: zero_cost_cond} for $(\q, \v) \in \mathbb{G}$. It should be noted that not all tasks are subject to the limitations imposed by underactuation, such as the swinging tracking of a fully actuated leg. The trajectory stabilization of a swinging leg can be accomplished using techniques analogous to those employed in the control of a fully actuated robot arm, which does not currently present a significant challenge.

To further find out what causes underactuation of floating-base motion, the state $(\q,\v)$ is separated into two parts, namely actuated joints state $(\q_a,\v_a)$ and unactuated floating-base state $(\q_{b},\v_b)$. As a result, the floating-base dynamics can be extracted from eq.~\eqref{eq: mrbd_dyn} as below \cite{kuindersma2016optimization}:
\begin{equation}\label{eq: floating-base dyn}
    \bM_b(\q)\dot \v_b + \bD_a(\q)\dot \v_a + \bb_b(\q,\v) =  \bJ_{cb}^T(\q)\bm{\lambda}.
\end{equation}
Once $(\q,\v)$ is given, $\dot\v_b$ only depends on the value of $\dot\v_a$ and $\bJ_{cb}^T\bm{\lambda}$. In the process of maintaining contact, namely let $(\q,\v)\in\mathbb{G}$, the following acceleration-level constraints can be derived:
\begin{equation} \label{eq: cstr_contact_acc}
    \frac{d}{dt}\big(\bJ_c(\q)\v\big) = \begin{bmatrix}
        \bJ_{cb}(\q) & \bJ_{ca}(\q)
    \end{bmatrix}\begin{bmatrix}
        \dot\v_b \\ \dot\v_a
    \end{bmatrix} + \dot\bJ_c(\q)\v = \bm{0}.
\end{equation}
Then we can solve $\dot\v_a$ from eq.~\eqref{eq: cstr_contact_acc}:
\begin{equation}
    \dot\v_a = -\bJ_{ca}^\dagger(\q)\big( \bJ_{cb}(\q)\dot\v_b + \dot\bJ_c(\q)\v\big)
\end{equation}
and substituting it back to eq.~\eqref{eq: floating-base dyn} yields
\begin{equation}\label{eq: floating-base dyn pure}
    \dot \v_b =  \hat\bM_b^{-1}(\q)\bJ_{cb}^T(\q)\bm{\lambda} - \hat\bM_b^{-1}(\q)\hat \bb_b(\q,\v)
\end{equation}
where 
\begin{subequations}
    \begin{align}
        \hat\bM_b(\q) &= \bM_b(\q) - \bD_a(\q)\bJ_{ca}^\dagger(\q)\bJ_{cb}(\q), \\
        \hat \bb_b(\q,\v) &= \bb_b(\q,\v)-\bJ_{ca}^\dagger(\q)\dot\bJ_c(\q)\v.
    \end{align}
\end{subequations}
At this point we can find that the value of $\dot \v_b$ is limited by the row space of $\bJ_{cb}^T(\q)$ and friction-cone constraints $\bm{\lambda}\in\mathbb{\hat S}$. Consequently, it is not feasible to simply transform the system \eqref{eq: floating-base dyn pure} into a double integrator system.


Now we choose Euler angle represents the pose of floating base and let $\q_b=\y_b\in\R^6$, then $\dot \q_b = \v_b$. Moreover, the linear feedback law given by eq.~\eqref{eq: tsfb_fb} is a way of implementing task-space feedback without considering the constraints imposed by the friction cone and the row-rank of $\bJ_{cb}^T(\q)$. It is precisely due to the neglect of these constraints that  the QP-based inverse dynamics solution is highly likely to fail in satisfying eq.~\eqref{eq: zero_cost_cond}. As a result, the use of user-tuned eq.~\eqref{eq: tsfb_fb} for task-space feedback yields poor performance.

Hence, our objective is to develop a methodology to construct task-space feedback for floating base that explicitly incorporates the aforementioned constraints, allowing the designed floating-base feedback to achieve better performance than the user-tuned approach.

\section{Riccati-based Feedback for Floating-Base}

In this section we will elaborate on how to construct a reasonable feedback control law for floating base that can explicitly consider the system underactuation and the friction-cone constraints. The key idea is to formulate this problem as a LQR by approximating the unactuated dynamics using a linear model and converting the friction-cone constraints to a soft one. Subsequently, a Riccati gain can be computed, and it is leveraged to establish a Riccati-based feedback control strategy for the floating-base task. Distinguishing our approach from existing methods, we refrain from direct utilization of the optimal trajectory or value function obtained from the LQR. Instead, we demonstrate the integration of the Riccati gain within WBC to achieve enhanced performance.


\subsection{Linear Model around Feasible Trajectory Reference}
Firstly, a linear model is employed to approximate the nonlinear unactuated dynamics. The following paragraph shows how to obtain such a linear model.

Due to the restriction on maintaining contact, namely $(\q,\v)\in\mathbb{G}$, noncollocated partial feedback linearization (section 3.5 in \cite{olfati2001nonlinear}) can not be applied to this system. Suppose the trajectory reference $(\q^*(t),\v^*(t))$ obtained by high-level planning is given. Substitute them into eq.~\eqref{eq: floating-base dyn pure} and then obtain the following linear model
\begin{equation}\label{eq: linear_model}
    \dot \v_b =  \underbrace{\hat\bM_b^{-1}(\q^*)\bJ_{cb}^T(\q^*)}_{\bB_\lambda}\bm{\lambda} - \underbrace{\hat\bM_b^{-1}(\q^*)\hat \bb_b(\q^*,\v^*)}_{\bc}
\end{equation}
with the friction-cone constraint $\bm{\lambda}\in\mathbb{\hat S}$ which is linear and convex. As we have chosen $\q_b = \y_b$ and now let $\bx_k = (\y_b(k\Delta t), \dot\y_b(k\Delta t))$, then the discrete-time linear dynamics for LQR is written as
\begin{align}
    \bx_{k+1} &= \underbrace{\begin{bmatrix}
        \bm{0} & \Delta t\cdot\bm{1} \\
        \bm{0} & \bm{0} 
    \end{bmatrix}}_{\bA}\bx_k  + \underbrace{\begin{bmatrix}
        \bm{0} \\ \Delta t\cdot\bB_\lambda(k\Delta t)
    \end{bmatrix}}_{\bB_k}\bm{\lambda}_k \nonumber\\
    & + \underbrace{\begin{bmatrix}
        \bm{0} \\ - \Delta t\cdot\bc(k\Delta t)
    \end{bmatrix}}_{\bd_k}
\end{align}
where $\Delta t$ is discretization time step.

\subsection{Handle Friction-Cone Constraints}
In this subsection, we will show how to handle the friction-cone constraints or how to consider the impact of this constraint in the final LQR problem. 


Now define the following quadratic cost function over a finite horizon $N$ steps
\begin{align}
    \J(\x_k,\bU_k, k) = \e_{k+N}^T\bP\e_{k+N} + \sum_{i=k}^{K+N-1}\e_i^T\bQ\e_i + \bm{\lambda}^T_{i}\bR\bm{\lambda}_{i}
\end{align}
where $U_k=\{\bm{\lambda}_k, \bm{\lambda}_{k+1}, ...,\bm{\lambda}_{k+N-1}\}$, $\e_i = \x_{i} - \x_{i}^*$ and $\x_k$ is the state at time $k\Delta t$. Consider the finite time optimal control problem
        \begin{subequations}\label{opt: constrained_LQR}
            \begin{align}\J^*(\x_k, k)  &=
                \min_{\bU_k}\quad \J(\x_k,\bU_k, k) \\
                \text{subj. to}\quad& \x_{i+1} = \bA\x_i + \bB_i\bm{\lambda}_i+\bd_i,\ \bm{\lambda}_i\in\mathbb{\hat S}, \\
                 &i=k,k+1,...,k+N-1.
            \end{align}
        \end{subequations}
Problem \eqref{opt: constrained_LQR} is a multiparametric quadratic program where the initial state $\x_k$ and time step $k$ are considered as the parameter set. Upon successfully solving this multiparametric program, the resulting solution, $\bar{\bU}_k(\x_k)$, for \eqref{opt: constrained_LQR} can be explicitly obtained as a function of the initial state. The approach outlined in \cite{borrelli2017predictive} (section 6.3) provides a methodology to tackle such multiparametric problems effectively. However, it is important to note that due to the considerable computational time required to solve such problems, they do not meet the real-time demands typically associated with the WBC which commonly operates at a frequency of 1 kHz. Consequently, this approach is not considered practical for real-time implementation. 

Although it is difficult to solve the global feedback law from the above multiparametric program in real time, it is possible to compute a local feedback law for given state $\x_k$ at time step $k$. Because \eqref{opt: constrained_LQR} degenerates into a quadratic programming when $\x_k$ and $k$ are fixed. Hence, we adopt a receding horizon control strategy, wherein we solve the quadratic program associated with a given $\x_k$ at each time step $k$. Then we can get the numerical value of $\bar{\bU}_k$ and the corresponding state trajectory $\{\bar\x_k^,\bar\x_{k+1},...,\bar\x_{k+N}\}$. This pipeline can be run in $50-100$Hz.

Now let $\Delta \bm{\lambda}_i = \bm{\lambda}_i - \bm{\bar\lambda}_i$. Here we choose logarithmic barrier function, a widely used barrier function used in interior-point methods, to represent the friction-cone constraints:
\begin{equation}
    \bC\bm{\lambda}_i\le \bn \Rightarrow \bm{\beta}(\Delta \bm{\lambda}_i) = \ln(\bn-\bC(\Delta \bm{\lambda}_i + \bm{\bar\lambda}_i)).
\end{equation}
Then we take second-order Taylor expansion $\bm{\hat\beta}(\Delta \bm{\lambda}_i)$ at $\Delta \bm{\lambda}_i = \bm{0}$ and append it to the running cost:
\begin{equation}
    \hat\J(\x_k,\bU_k, k) = \J(\x_k,\bU_k, k) + \sum_{i=k}^{k+N-1}\bm{\hat\beta}(\Delta \bm{\lambda}_{i}).
\end{equation}
Now we can write out the resulting LQR to compute a task-space Riccati feedback gain:
\begin{subequations}\label{opt: constrained_LQR_approx}
    \begin{align}\hat\J^*  &=
        \min_{\Delta\bU_k}\quad \hat\J(\bar\x_k+\Delta\x_k,\bar\bU_k+\Delta\bU_k, k) \\
        \text{subj. to}\quad& \Delta \x_{i+1} = \bA\Delta\x_i + \bB_i\Delta\bm{\lambda}_i,\\
         &i=k,k+1,...,k+N-1,\\
         &\Delta\x_k = \x(k\Delta t) - \bar\x_k.
    \end{align}
\end{subequations}
where $\Delta\x_i=\x_i-\bar\x_i$ denotes the difference between system trajectory and its reference. Now by solving the time-variant LQR we can obtain the following feedback law for every time step:
\begin{equation}
    \Delta\bm{\lambda}_k = \bF^{lqr}_k\Delta\x_k \Rightarrow  \bm{\lambda}_k = \bm{\bar\lambda}_k + \bF^{lqr}_k\Delta\x_k
\end{equation}
and substituting it back to \eqref{eq: linear_model} yields
\begin{equation}
    \ba_b = \dot \v_b(t) = \bB_\lambda(t)(\bm{\bar\lambda}_k + \bF^{lqr}_k\Delta\x(t)) - \bc(t)
\end{equation}
for any $t\in [k\Delta t, (k+1)\Delta t\big)$. 

Compared to the user-tuned feedback law \eqref{eq: tsfb_fb}, user-tuned parameters $\bK_p^b,\bK_d^b$ are replaced by a time-variant gain $\bB_\lambda(t)\bF^{lqr}_k$ and the feedforward term is replaced by $\bB_\lambda(t)\bm{\bar\lambda}_k-\bc(t)$:
\begin{subequations}
\begin{align}
    \begin{bmatrix}
        \bK_p^b & \bK_d^b
    \end{bmatrix} &\leftarrow \bB_\lambda(t)\bF^{lqr}_k \\
    \ddot \y_b^*(t) &\leftarrow \bB_\lambda(t)\bm{\bar\lambda}_k-\bc(t)
\end{align}
\end{subequations}

As mentioned before, the feasible domain of $\ba_b$ is limited by the row space of $\bJ_{cb}^T(\q)$ and friction-cone constraints. Now these factors have been taken into consideration in the resulting LQR to compute the Riccati gain $\bF^{lqr}$. Instead of tuning the parameters heuristically, the proposed method computes the feedback law in a reasonable way with more model information. Based on this point, the proposed method should achieve better control performance.

\section{Validation}
\begin{figure}[h]
	\centering
 \captionsetup[subfloat]{labelfont=scriptsize,textfont=scriptsize}
	\subfloat[]{\includegraphics[width=.15\linewidth]{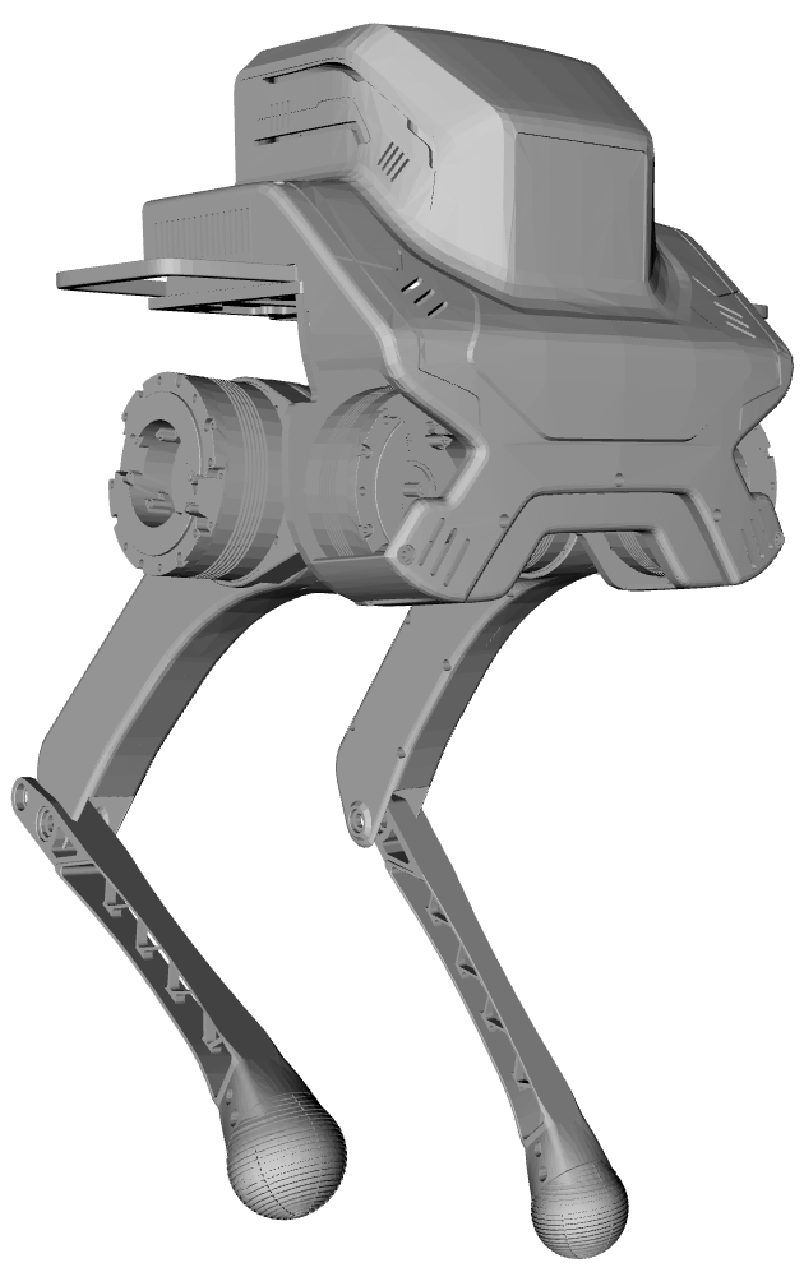}}
 \hspace{1cm}
	\subfloat[]{\includegraphics[width=.3\linewidth]{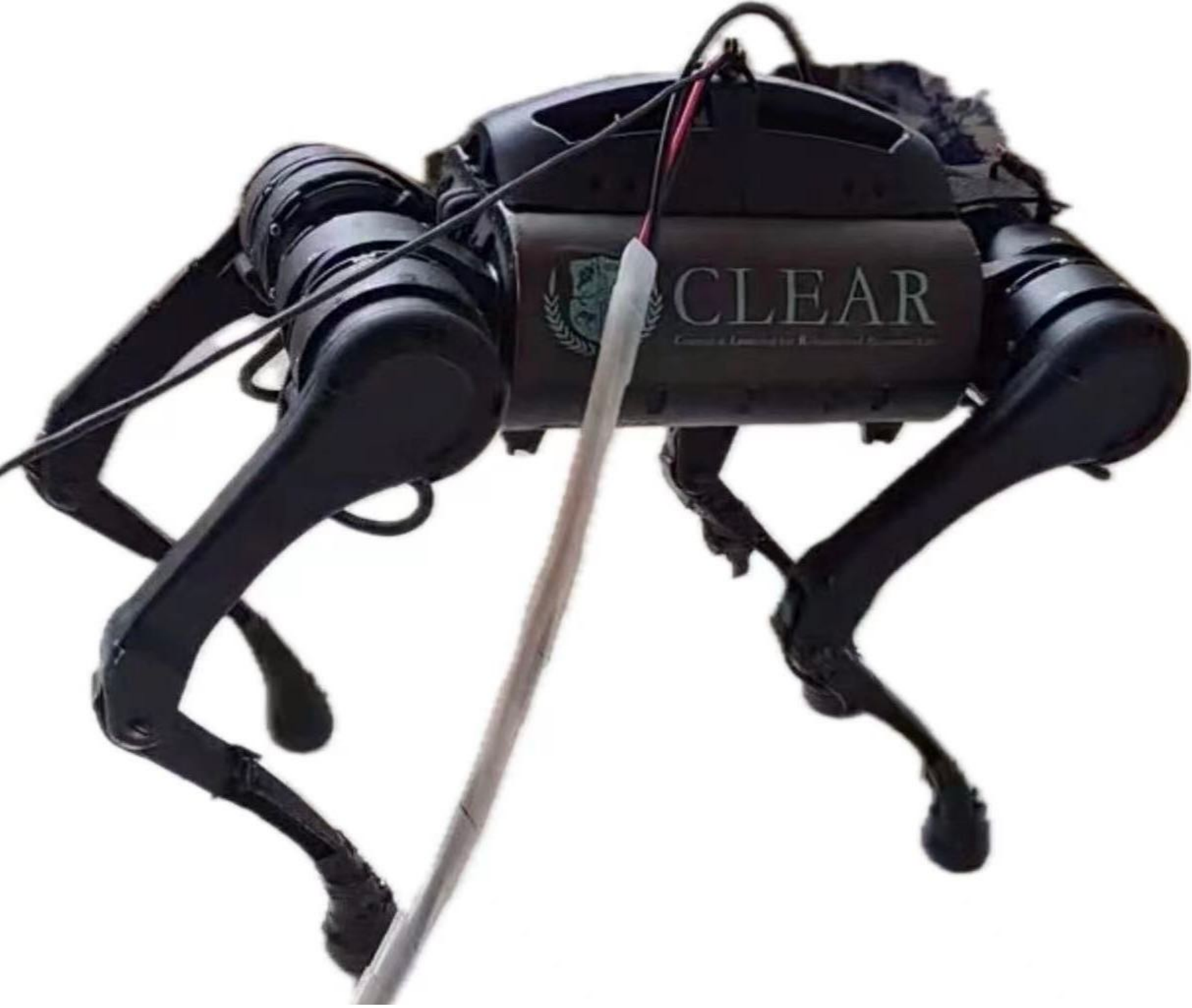}}\\
	\caption{(a) Bipedal robot P1 used for simulation tests (b) Quadruped robot A1 for experiments.}
    \label{fig: robots}
    \vspace{-0.5cm}
\end{figure}

To validate the effectiveness of the proposed method, some experiments are conducted to compare the proposed method with user-tuned method. Furthermore, two typical legged robots shown as Fig.~\ref{fig: robots}, Unitree A1 (quadruped) and LimX Dynamics P1 (biped), are used in the tests. The whole project is implemented in C++ and based on ROS2 Humble due to their high efficiency and easy deployment on real robot. Besides, HPIPM\cite{frison2020hpipm} is adopted to solve LQR due to its high computational efficiency.
\vspace{-0.3cm}
\begin{figure}[h]
    \centering
    \includegraphics[width=0.47\textwidth]{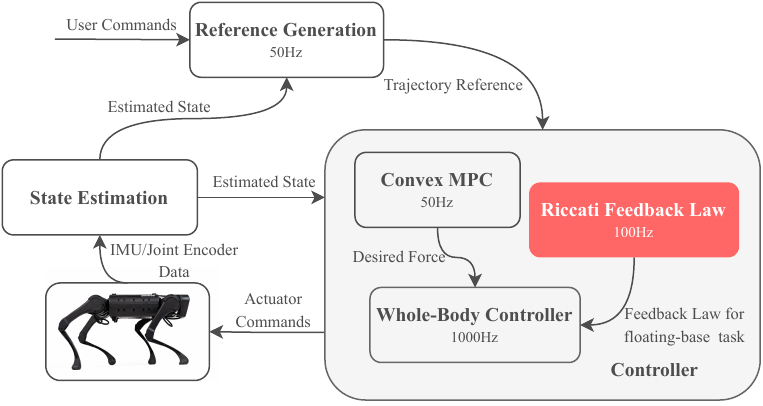}
    \caption{Framework for robot control encompassing state estimation, reference generation(planning), and low-level controller. The control module integrates convex MPC, LQR in eq.~\eqref{opt: constrained_LQR_approx} and whole-body controller.}
    \label{fig: framework}
    \vspace{-0.5cm}
\end{figure}

\subsection{Trajectory Reference Generation}
 For bipedal robot P1, a linear inverted pendulum model\cite{kajita20013d} is used to generate floating-base motion reference and determine footholds. For quadruped A1, footholds are selected using the Raibert heuristic method\cite{raibert1986legged}, while the swinging trajectory reference is fitted using cubic spline interpolation. Additionally, the floating-base trajectory reference is obtained by a simple double integrator. Then the trajectory reference in joint space can be obtained by solving inverse kinematics, which is used to compute the linear model. Furthermore, the force reference for force task in WBC is generated by convex MPC\cite{di2018dynamic} in 50Hz.

\subsection{Bipedal Simulation Tests}
In order to validate the efficacy of the novel algorithm on bipedal robots, a total of six tests were conducted in the simulation, with the key distinction between the two tests lying in the variance of the upper-level velocity commands. The tests encompassed movements ranging from unidirectional to omnidirectional, providing a comprehensive evaluation.

In a series of tests, the increment in the instruction was initially applied solely in the x-direction. \textbf{Test 1} involved setting the x-direction velocity instruction to $0.3$ m/s, \textbf{Test 2} set it to $0.6$ m/s, and \textbf{Test 3} set it to $0.9$ m/s. Subsequently, the velocity instruction was individually increased in the y-direction while keeping the remaining velocity instructions at zero. In \textbf{Test 4} and test \textbf{Test 5}, the y-direction velocity instructions were set to $0.2$ m/s and $0.3$ m/s, respectively. In the final \textbf{Test 6}, the velocities in all directions simultaneously followed a sinusoidal trend. The specific values were as follows: 
$v_x = 0.3\sin(t)$ m/s, $v_y = 0.3\sin(t)$ m/s, and $v_{yaw} = 0.4\sin(2t)$ rad/s.
\vspace{-0.3cm}
\begin{table}[h]
\caption{\textbf{Biped} Mean Square Error (Linear Velocity, Unit: $m^2/s^2$)}
\label{table: lin_mse_bipedal}
\vspace{-0.5cm}
\begin{center}
\begin{tabular}{|c||c|c|c|c|}
\hline
 & User-Tuned 1 & User-Tuned 2 & User-Tuned 3 & \textbf{Proposed} \\
\hline
Test 1 & 0.0420 (19\%)  & 0.0393 (13\%)& 0.0458 (26\%) & \textbf{0.0338} \\
\hline
Test 2 & 0.0499 (27\%)& 0.0499 (27\%)& 0.0565 (36\%)& \textbf{0.0361} \\
\hline
Test 3 & 0.0582 (35\%)& 0.0554 (31\%)& 0.0651 (42\%)& \textbf{0.0377} \\
\hline
Test 4 & 0.0352 (17\%)& 0.0349 (16\%)& 0.0373 (21\%)& \textbf{0.0291} \\
\hline
Test 5& 0.0345 (\textbf{12\%})& 0.0353 (14\%)& 0.0386 (22\%)& \textbf{0.0301} \\
\hline
Test 6 & 0.0558 (31\%)& 0.0450 (15\%)& 0.0763 (\textbf{50\%})& \textbf{0.0380} \\
\hline
\end{tabular}
\end{center}
\vspace{-0.3cm}
\end{table}
\vspace{-0.3cm}
\begin{table}[h]
\caption{\textbf{Biped} Mean Square Error (Angular Velocity, Unit: $rad^2/s^2$)}
\label{table: ang_mse_bipedal}
\vspace{-0.5cm}
\begin{center}
\begin{tabular}{|c||c|c|c|c|}
\hline
 & User-Tuned 1 & User-Tuned 2 & User-Tuned 3 & \textbf{Proposed} \\
\hline
Test 1 & 0.5355  (27\%)& 0.5367 (27\%)& 0.6599  (40\%)& \textbf{0.3907} \\
\hline
Test 2 & 1.0741 (31\%)& 1.1303 (35\%)& 1.3978 (\textbf{47\%})& \textbf{0.7329} \\
\hline
Test 3 & 1.5580 (37\%)& 1.5356 (36\%)& 1.3707 (28\%)& \textbf{0.9737} \\
\hline
Test 4 & 0.1896 (18\%)& 0.1771 (12\%)& 0.1677 (\textbf{7\%})& \textbf{0.1546} \\
\hline
Test 5& 0.1907 (\textbf{7\%})& 0.2031 (13\%)& 0.2126 (17\%)& \textbf{0.1762} \\
\hline
Test 6 & 0.5565 (20\%)& 0.6065 (26)& 0.7401 (40\%)& \textbf{0.4430} \\
\hline
\end{tabular}
\end{center}
\vspace{-0.5cm}
\end{table}

In the case of the conventional approach, three sets of different parameters were employed, which were not arbitrarily chosen but rather selected after experimenting with multiple parameter combinations to identify the best-performing ones. However, despite these efforts, it was challenging for the parameter tuning method in WBC to achieve the same level of performance as those obtained by the new approach, as evidenced by the velocity tracking errors in Table~\ref{table: lin_mse_bipedal} and Table~\ref{table: ang_mse_bipedal}. In all conducted tests, the newly proposed method consistently exhibits the lowest mean squared error in terms of velocity. While certain user-tuned parameters in a specific test (such as \textbf{Test 4} and \textbf{Test 5}) may approach the results of the new method, overall, fixed parameters fail to achieve the effectiveness of the new method across all tests.


\begin{figure}[t!]
    \centering
    \includegraphics[width=0.5\textwidth]{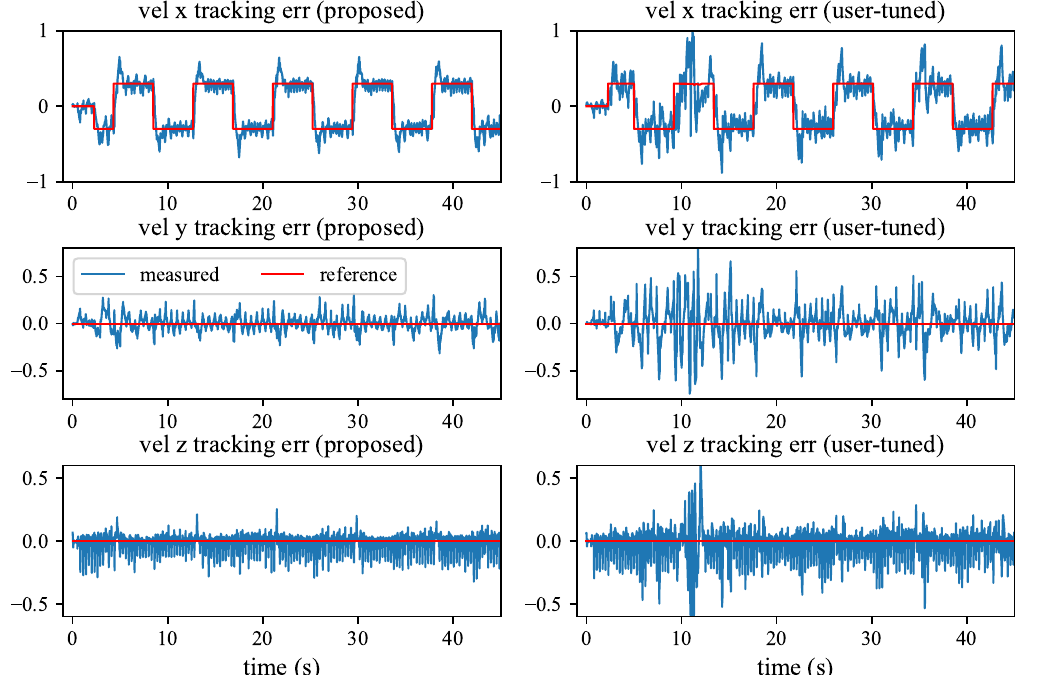}
    \caption{Linear velocity in quadruped hardware Test 7 (Blue: measured, Red: reference).}
    \label{fig: quad_linear_err_test1}
\end{figure}
\begin{figure}[t!]
    \centering
    \includegraphics[width=0.5\textwidth]{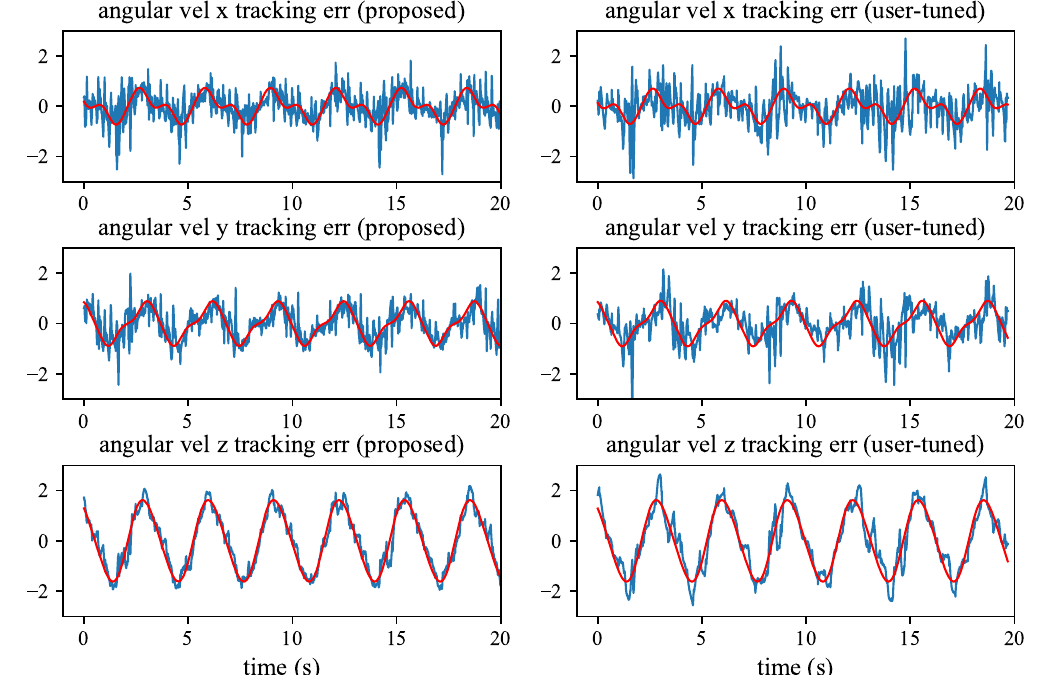}
    \caption{Angular velocity in quadruped hardware Test 9 (Blue: measured, Red: reference).}
    \label{fig: quad_ang_err_test3}
    \vspace{-0.5cm}
\end{figure}
\subsection{Quadruped Experiments}
To further validate the proposed algorithm on a physical platform, we employed a commercially available quadruped robot, namely A1. In the bipedal simulation tests, we have demonstrated the advantages of a novel approach in tracking smooth velocity commands. In the forthcoming quadruped robot experiments, we aim to further compare the performance of the new and conventional methods when confronted with non-smooth velocity commands. As depicted by the red line in Fig.~\ref{fig: quad_linear_err_test1}, the direction of the instruction undergoes sudden changes, transitioning abruptly from forward motion to backward motion. In \textbf{Test 7}, the magnitude of the velocity instruction in the x-direction was set to $0.3$ m/s, but it varied between forward and backward motion. After completing the x-direction test, \textbf{Test 8} focused on the y-direction with a velocity magnitude of $0.2$ m/s. In \textbf{Test 9}, the objective was to track a smooth velocity instruction following a sinusoidal trend. The specific values for this instruction were as follows: $v_x = 0.3\sin(2t)$ m/s, $v_y = 0.4\sin(2t)$ m/s, and $v_{yaw} = 0.8\sin(2t)$ rad/s.

During the experiments, we observed that the parameters $\bK_p^b, \bK_d^b$ in the user-tuned method were set too large, resulting in oscillations of the robot's floating base. After extensive tuning, we selected the proportional gain $\bK_p^b$ and derivative gain $\bK_d^b$ as diagonal matrices, where the values on the diagonal were set to 20 and 1, respectively.
Furthermore, in addition to setting the weights of floating-base task and swinging task to 100, assigning a weight between 10 and 50 to force task can also attenuate floating base oscillations and yield improved control performance in the experiments.

During the deployment process, we observed that the parameters tuned for the old method in simulation required readjustment when deployed on hardware. In contrast, the new method did not necessitate such readjustment. This finding further highlights that the user-tuned method demands significant effort in parameter tuning, whereas the new method, to some extend, reduces the effort spent on parameter tuning.
\vspace{-0.3cm}
\begin{table}[h]
\caption{\textbf{Quadruped} Mean Square Error (Unit: $m^2/s^2, rad^2/s^2$)}
\label{table: mse_quad}
\vspace{-0.5cm}
\begin{center}
\begin{tabular}{|c||c|c|c|}
\hline
 & User-Tuned ($v$, $\omega$) & \textbf{Proposed} ($v$, $\omega$) & Improvement \\
\hline
Test 7 & (0.2036, 0.6676)   & (\textbf{0.1246}, \textbf{0.3596}) & (38\%, 46\%) \\
\hline
Test 8 & (0.1572, 0.5849) & (\textbf{0.1518}, \textbf{0.4295}) & (3\%, 26\%)\\
\hline
Test 9 & (0.1795, 0.7148) & (\textbf{0.1697}, \textbf{0.5498}) & (5\%, 23\%)\\
\hline
\end{tabular}
\end{center}
\end{table}
\vspace{-0.3cm}

In the three conducted tests, as shown in Table~\ref{table: mse_quad}, the new method exhibits smaller velocity tracking errors. The improvement is particularly obvious in angular velocity tracking, reaching at least 23\%. Additionally, as evident from Fig.~\ref{fig: quad_linear_err_test1} and Fig.~\ref{fig: quad_ang_err_test3}, the proposed method demonstrates not only improved stability during smooth velocity tracking but also reduced oscillation amplitudes when faced with abrupt velocity instructions change.


    
    

\section{conclusion and discussion}
To address the issue of user-tuned methods requiring significant effort in parameter tuning with limited effectiveness, we explored the use of LQR to compute a Riccati gain and construct a feedback control law in the task space of floating base. Furthermore, we incorporated model information and physical constraints into the LQR formulation. In a sense, the traditional method relies on manual parameter tuning and is considered a model-free approach, while the new method adopts a model-based approach for parameter setting. Through simulations and experiments, we validated the positive impact of the new method on the performance improvement of WBC for legged robots. Additionally, the new method significantly reduces the effort and time spent on parameter tuning for floating-base task. Indeed, one limitation of the proposed method is that the linear model used in the LQR is a linearization of the nonlinear model around a specific trajectory. Consequently, if the robot's state deviates significantly from that trajectory, the effectiveness of the control law derived from this method becomes uncertain. The linearized model assumes a small range of deviation from the nominal trajectory, and beyond that range, its validity diminishes. On the other hand, the limitations can be mitigated to some extent by using replanning techniques, especially when combined with mature motion planning algorithms in the higher-level trajectory optimization.

\bibliographystyle{IEEEtran}
\bibliography{ref}

\end{document}